Posterior-Calibrated Causal Circuits in Variational Autoencoders: Why Image-Domain Interpretability Fails on Tabular Data

# Posterior-Calibrated Causal Circuits in Variational Autoencoders: Why Image-Domain Interpretability Fails on Tabular Data

Dip Roy[1*], Rajiv Misra[1], Sanjay Kumar Singh[2], Anisha Roy[3]

[1]*Department of Computer Science and Engineering, Indian Institute of Technology, Patna, India*

[2]*Department of Computer Science, Rajarshi School of Management Technology, Varanasi, India*

[3] *Department of Electronics and Communication Engineering, Jaypee Institute of Information Technology, Noida*

*Corresponding author: dip_25s21res37@iitp.ac.in

## Abstract

Although mechanism-based interpretability has generated an abundance of insight for discriminative network analysis, generative models are less understood — particularly outside of image-related applications. We investigate how much of the causal circuitry found within image-related variational autoencoders (VAEs) will generalize to tabular data, as VAEs are increasingly used for imputation, anomaly detection, and synthetic data generation. In addition to extending a four-level causal intervention framework to four tabular and one image benchmark across five different VAE architectures (with 75 individual training runs per architecture and three random seed values for each run), this paper introduces three new techniques: posterior-calibration of Causal Effect Strength (CES), path-specific activation patching, and Feature-Group Disentanglement (FGD). The results from our experiments demonstrate that: (i) Tabular VAEs have circuits with modularity that is approximately 50% lower than their image counterparts. (ii) β-VAE experiences nearly complete collapse in CES scores when applied to heterogeneous tabular features (0.043 CES score for tabular data compared to 0.133 CES score for images), which can be directly attributed to reconstruction quality degradation ($r = -0.886$ correlation coefficient between CES and MSE). (iii) CES successfully captures nine of eleven statistically significant architecture differences using Holm–Šídák corrections. (iv) Interventions with high specificity predict the highest downstream AUC values ($r = 0.460$, $p < .001$). This study challenges the common assumption that architectural guidance from image-related studies can be transferred to tabular datasets.

## 1. Introduction

Although they have been used for many years in safety-critical domains, most deep neural networks are still essentially "black boxes." Mechanistic interpretability seeks to turn those black boxes into transparent systems by reverse engineering the internal circuitry that generates a model's outputs. Much progress toward this goal has been made with discriminative models through landmark discoveries (e.g., induction heads in language models [1], indirect object identification circuits in GPT-2 [2], and interpretable feature hierarchies in convolutional networks [3]), but much less research has focused on how generative models organize their internal computation.

Generative models present fundamentally different interpretability challenges. Variational autoencoders (VAEs) [4] learn to map high-dimensional observations into structured latent spaces and reconstruct the original data, introducing additional layers of representational complexity through the encoder-decoder architecture and the stochastic latent bottleneck. Several architectural variants—including β-VAE [5], FactorVAE [6], β-TC-VAE [7], and DIP-VAE [8]—have been proposed to encourage disentangled representations. While existing disentanglement metrics such as DCI [9], MIG [7], and SAP [8] evaluate the quality of these learned representations, they treat the model as a black box: they measure input-output relationships without revealing the internal mechanisms through which disentanglement emerges or fails.

Roy and Misra [11] recently introduced a multi-level causal intervention framework that addresses this gap for image-domain VAEs. By performing interventions at four levels of the network—input modifications, latent space perturbations, activation patching, and causal mediation analysis—they exposed the internal circuits of six VAE architectures across five image benchmarks, uncovering phenomena such as a CES–DCI tradeoff and a β-VAE capacity bottleneck. However, a critical assumption underlies this and virtually all prior MI work on VAEs: that the mechanistic phenomena discovered on image data represent general properties of VAE computation, rather than artifacts of the image domain.

This assumption should be carefully considered because VAEs are being used increasingly outside of images. In addition to images, tabular applications – including missing data imputation [12], anomaly detection [13] and generating synthetic data [14] -- have become significant uses. Images differ from tabular data in many key ways: images include both discrete (categorical) and continuous data; they do not possess spatial locality for convolutional layers to utilize; and the semantic meaning of each image pixel exists as part of a uniform grid structure while the semantic meaning of each data record exists within a domain specific grouping. Therefore, it is important to understand if the circuitry that VAEs have learned to process tabular data in the same way that VAEs have learned to process images will allow the increasing number of image-domain MI results to help inform practitioners using tabular VAEs.

This paper addresses this gap through a systematic cross-modality study. Specifically, we investigate four research questions:

**RQ1:** Do the causal circuit structures identified by MI techniques in image-domain VAEs transfer to the tabular domain? If not, how do the circuits differ in terms of modularity, specificity, and causal effect strength?

**RQ2:** Is the β-VAE capacity bottleneck—where strong KL reweighting suppresses per-dimension representational strength—modality-dependent?

**RQ3:** Which causal intervention metrics are most effective at discriminating between VAE architectures, and does this hold across modalities?

**RQ4:** Do the mechanistic circuit properties predict downstream task performance on tabular data, and if so, which properties matter most?

Our contributions are fivefold. First, we present the first systematic cross-modality study of VAE mechanistic interpretability, evaluating whether image-domain findings generalize to tabular data through 75 independent training runs. Second, we introduce three methodological refinements—posterior-calibrated CES, path-specific activation patching, and Feature-Group Disentanglement (FGD)—each accompanied by formal theoretical justifications establishing their validity and relationship to existing metrics. Third, we establish that the β-VAE capacity bottleneck is modality-dependent and operates through reconstruction degradation, not merely latent space reshaping. Fourth, we demonstrate that CES is the most architecturally discriminative causal metric and that intervention specificity is the strongest predictor of downstream quality for tabular VAEs. Fifth, we identify a structural limitation of causal mediation analysis in sequential architectures that was not apparent in prior convolutional-encoder studies.

## 2. Related Work

### 2.1 Mechanistic Interpretability

Mechanistic interpretability seeks to understand neural computation by identifying the circuits—subsets of model components and their connections—that implement specific functions. Olah et al. [3] demonstrated that individual neurons and groups of neurons in convolutional networks form interpretable features organized into circuits. Elhage et al. [1] developed a mathematical framework for analyzing attention-based circuits in transformers, which subsequent work has used to identify mechanisms such as indirect object identification in GPT-2 [2] and factual association storage [15]. Network Dissection [16] provides quantitative methods for measuring semantic correspondence of neurons. Recent work on sparse autoencoders [34, 35] has extended mechanistic interpretability by training auxiliary models to decompose activations into interpretable features, providing evidence that neural networks represent information in superposition.

Most of the literature in this area uses discrimination-based architectures. The application of generative architectures presents many different challenges. First, the model's output is a high dimensional reconstruction rather than a discrete prediction. Second, the stochastic nature of the latent space adds another dimension to the representation. Bau et al. [17] applied network dissection to GANs, identifying neurons that controlled specific visual attributes. Roy et al. [11], provided the first general causal intervention framework for VAEs; they performed interventions at four levels on six architectures using five image data sets (a total of 90 training runs). In addition to CES, the authors also proposed the use of specificity, and circuit modularity as quantitative measures. Our paper builds upon their framework by extending it to the table based domain with added formal theoretical refinement.

### 2.2 Disentangled Representation Learning

The motivation behind pursuing "disentangled" representations in VAEs has generated many variants of the VAE. Higgins et al. [5], for example, proposed β-VAE by increasing the weight of the KL divergence loss so as to encourage factorial posteriors. Kim and Mnih [6], proposed FactorVAE, and used an adversarial discriminator to penalize the total correlation (TC) of the aggregate posterior. Chen et al. [7], developed β-TC-VAE, that separates the KL loss into three distinct terms: mutual information, total correlation, and dimension-wise KL losses. Kumar et al. [8], proposed DIP-VAE, that adds regularization on the covariance of the aggregate posterior to make it similar to a factorized prior.

The authors of [19] have done an excellent job to show that theoretical disentanglement from unsupervised data cannot be achieved without some form of inductive bias. In addition, they demonstrated that no single technique for achieving disentanglement is superior when compared on all possible datasets with various hyperparameter configurations. They advocated for multi-seed testing and a broad suite of evaluation metrics, which are both methodologies we also utilize. The standard metrics used for evaluating representations through their relationship to input-output statistics include DCI [9], MIG [7], SAP [8], β-metric [5], and FactorVAE metric [6]. Do and Tran [20] examined the theoretical basis of each of these metrics, and found that each measures a unique aspect of the representation space. We provide additional mechanistic explanation.

## 2.3 Causal Methods in Neural Network Analysis

The use of Pearl's causal inference framework [21] for neural networks is based on the idea that correlation does not equal cause – establishing causality typically involves intervening. A bridge between the two was created by Geiger et al. [22], who formalized causal abstraction. Meng et al. [15] provided an example of causal tracing with very large language models. Patching activations have been shown to be useful for identifying causal paths [23]. Vig et al. [24] applied causal mediation analysis to examine gender bias in NLP models. Yang et al. [25] introduced CausalVAE which integrates causal structure into the generative mechanism. Suter et al. [26] examined robustly disentangled causal mechanisms. While our method is different from CausalVAE as it can be applied after training to any VAE.

## 2.4 VAEs for Tabular Data

While VAEs were initially designed for images, there has been a significant increase in VAE use for tables. Xu et al. [14] showed that TVAE could be used for generating synthetic table data with similar or better results than other approaches. Ma et al. [27] utilized VAEs to predict missing values in clinical data sets. VAE-based anomaly detection using the reconstruction loss function has also been implemented in various industrial [13] and finance sectors. Other examples of tabular deep learning include architectures like TabNet [36], SAINT [37], and FT-Transformer [38] that are comparable to gradient boosting algorithms. The internal mechanisms of tabular VAEs—i.e., how they represent different types of heterogeneous features, what circuitry forms within them, and if those resemble the circuitry formed in the domain of images—have yet to be explored.

## 2.5 Information Bottleneck Theory

The Information Bottleneck Principle [39], gives a theoretical basis to understand the β-VAE capacity bottleneck identified in the paper. In their work, Tishby et al. [39] defined the relationship of the trade-off between the ability to compress data and to predict it. Alemi et al. [40] also showed that the VAE loss function (the objective function) is related to the information bottleneck principle. They showed that an increase in the β parameter would tighten the compression constraint. As such, we can view our finding that β-VAEs collapse much more severely when trained with tabular data, as being due to the fact that each feature in tabular data has significantly more "information" in terms of the number of dimensions than does each pixel neighborhood, resulting in the bottleneck being more severe for the same value of β.

## 2.6 Positioning of This Work

Our study is an innovative intersection of three broad areas: Mechanistic Interpretability [1, 3, 11], Disentangled Representation Learning [5, 6, 7, 8] and Tabular Deep Learning [14, 27, 36]. Roy et al. [11] have established Mechanistic Interpretation (MI) techniques for Image-Domain VAEs. We are investigating if they will apply. The literature on Disentanglement has primarily focused on image benchmarks [19]. We are testing if the same architecture trade-offs hold for Tabular Data. The literature on Tabular VAE's has focused primarily on the Quality of Generation. We are providing the first Mechanistic Analysis to understand how Tabular VAE's organize their computation.

## 3. Methodology

### 3.1 Problem Formulation

Consider a VAE with encoder $E_\phi$ and decoder $D_\theta$ trained on data drawn from either the image domain (where the generative process involves K independent factors of variation) or the tabular domain (where features form G semantically meaningful groups). For input $x \in \mathbb{R}^n$, the encoder produces approximate posterior parameters $q_\phi(z|x) = N(\mu_\phi(x), \mathrm{diag}(\sigma^2_\phi(x)))$, and the decoder reconstructs $\hat{x} = D_\theta(z)$. Our objective is to identify how causal information pathways differ between these two settings, and to determine whether the circuit structures observed in one domain provide reliable predictions about the other.

We formalize the comparison through the lens of causal interventions. A *circuit* in this context is defined as the set of latent dimensions $\{d_1, ..., d_k\}$ and their associated encoder pathways that mediate information flow from a specific feature group g to the decoder output. The *strength* of a circuit is measured by CES (the causal effect of perturbing the latent dimension on the output), while its *specificity* measures the degree to which it responds selectively to its associated feature group.

### 3.2 Multi-Level Causal Intervention Framework

We adopt the four-level causal intervention framework introduced by Roy and Misra [11] with three refinements motivated by the distinct properties of tabular data. Each level probes a different layer of the computation and answers a distinct mechanistic question.

#### *3.2.1 Level 1: Multi-Scale Input Interventions*

*Question: How does the encoder transform feature-group perturbations into latent shifts?*

For each semantic feature group $g \in \{1, ..., G\}$ and perturbation scale $s \in \{0.5, 1.0, 2.0\}$ standard deviations, we perturb the corresponding features while holding all others at their original values. The latent response is measured as:

$$\Delta z^{(s)}_g = E_x[\, \|E_\phi(\tilde{x}^{(s)}_g) - E_\phi(x)\| \,] \quad (1)$$

where $\tilde{x}^{(s)}_g$ is the input with features in group g replaced by $x_g + s \cdot \sigma_g$ (with $\sigma_g$ being the per-feature standard deviation computed from training data) and all other features unchanged. The multi-scale design enables detection of nonlinear response profiles. We compute a linearity score $R^2$ by regressing response magnitude on perturbation scale for each feature group. Across our experiments, linearity scores consistently exceed $R^2 > 0.96$, indicating that the encoder response is approximately linear within the tested range for the fully-connected architectures used.

The Level 1 response matrix $\Delta \in \mathbb{R}^{(D\times G)}$ (latent dimensions × feature groups), averaged across perturbation scales, serves as the importance matrix R for computing FGD and Modularity (Section 3.3).

### 3.2.2 Level 2: Posterior-Calibrated CES

*Question: How much causal influence does each latent dimension exert on the decoder output?*

The original CES formulation [11] sweeps each latent dimension across a fixed range [−3, +3]. When posterior distributions vary substantially across dimensions—which is common in VAEs with regularization penalties—this conflates genuine causal influence with distributional mismatch. A dimension with narrow posterior variance ($\sigma^2 \approx 0.01$) would be tested at values 300 standard deviations from its mean under the fixed-range protocol, producing artificially large output perturbations that reflect extrapolation artifacts rather than meaningful causal effects.

We introduce posterior-calibrated CES to correct this bias:

$$CES(d) = E_x\left[ (1/|V|) \sum_{v\in V} (1/n) \|D_\theta(\tilde{z}_{d,v}) - D_\theta(\mu)\|_1 \right] \quad (2)$$

where $\tilde{z}_{d,v}$ sets dimension d to $v = \mu_d + \sigma^{eff}_d \cdot t$ for t spanning [−3, +3] in standard deviation units across $|V| = 51$ uniformly spaced evaluation points, and the effective standard deviation is:

$$\sigma^{eff}_d = \max(\mathrm{std}(\mu_d), \sqrt{\exp(\overline{\log \sigma^2}_d)}) \quad (3)$$

The first term, $\mathrm{std}(\mu_d)$, captures the aggregate spread of the posterior means across the dataset. The second term, $\sqrt{\exp(\overline{\log \sigma^2}_d)}$, captures the mean per-sample posterior uncertainty. Taking the maximum ensures that we sweep over the larger of the two sources of variation, avoiding the pathological case where a dimension has tightly clustered means but high per-sample uncertainty (or vice versa).

**Proposition 1** (CES Calibration Correction). *Let $CES_{fixed}(d)$ denote the original CES computed with fixed sweep range [−a, +a], and let $CES_{cal}(d)$ denote the posterior-calibrated CES. If the decoder $D_\theta$ is locally linear with Jacobian $J_d = \partial D_\theta/\partial z_d$ evaluated at the posterior mean, then:*

$$CES_{fixed}(d) \approx (\|J_d\|_1/n) \cdot (a/2)$$

$$CES_{cal}(d) \approx (\|J_d\|_1/n) \cdot \sigma^{eff}_d \cdot (a/2)$$

*Thus $CES_{cal}(d)/CES_{fixed}(d) \approx \sigma^{eff}_d$, and for a well-regularized dimension with $\sigma^{eff}_d \ll 1$, the fixed-range CES overestimates causal influence by a factor of $1/\sigma^{eff}_d$.*

*Proof sketch.* For a locally linear decoder, the absolute reconstruction shift when perturbing dimension d by $\delta$ is $|J_d \cdot \delta| = |J_d| \cdot |\delta|$ (element-wise). Averaging over the sweep range [−a, +a]: $E_\delta[|\delta|] = a/2$. For the calibrated version, $\delta = \sigma^{eff}_d \cdot t$ with $t \in [-a, +a]$, giving $E_t[|\sigma^{eff}_d \cdot t|] = \sigma^{eff}_d \cdot a/2$. The ratio follows. □

This result has practical significance: for a β-VAE dimension with $\sigma^{eff}_d \approx 0.1$ (common under strong KL regularization), the fixed-range CES overestimates influence by 10×, potentially masking the capacity bottleneck we aim to detect.

### 3.2.3 Level 3: Path-Specific Activation Patching

*Question: How much does each encoder layer contribute to the latent representation?*

Standard activation patching at layer l measures the compound effect of replacing activations at that layer plus all downstream processing. Specifically, given a source input $x_s$ and a target input $x_t$, compound patching replaces the activation at layer l with the source's activation and allows the modified signal to propagate through all subsequent layers:

$$\text{compound}(l) = \|\mu_\text{patched}(l) - \mu_t\| \quad (4)$$

where $\mu_\text{patched}(l)$ is the mean latent vector obtained when layer l receives source activations but subsequent layers process normally. We decompose this into direct contributions:

$$\text{direct}(l) = \text{compound}(l) - \text{compound}(l+1) \quad (5)$$

where compound(L) = 0 for the final encoder layer L (since patching after the last layer has no effect on the mean computation).

**Proposition 2** (Exact Decomposition in Sequential Architectures). *For a sequential encoder with layers* $l_0, l_1, \ldots, l_L$ *without skip connections or residual paths, the direct contributions form an exact telescoping decomposition:*

$$\Sigma_{l=0}^{L} \text{direct}(l) = \text{compound}(0)$$

*i.e., the sum of all direct contributions equals the total patching effect.*

*Proof.* By telescoping: $\Sigma_{l=0}^{L} [\text{compound}(l) - \text{compound}(l+1)] = \text{compound}(0) - \text{compound}(L+1) = \text{compound}(0)$, since $\text{compound}(L+1) = 0$ (no patching after the network). □

This decomposition is exact for the sequential fully-connected architectures in our study. For architectures with skip connections or residual blocks, the direct contributions would not sum exactly to compound(0) due to interaction terms, and the residual would quantify the strength of non-sequential pathways.

#### 3.2.4 Level 4: Clamped Causal Mediation with NIS

*Question: What fraction of a feature group's total effect is mediated through each encoder layer?*

For feature group g and encoder layer l, we compute the mediation ratio:

$$\text{MR}(g, l) = (\text{TE}(g) - \text{RE}(g, l)) / \text{TE}(g) \quad (6)$$

where TE(g) is the total effect of perturbing feature group g on the latent representation (measured as $\|\mu(\tilde{x}_g) - \mu(x)\|$), and RE(g, l) is the remaining effect when layer l is frozen to its clean-input activations (i.e., the activations computed from the unperturbed input x). Freezing layer l blocks the perturbed signal from propagating through that layer, so RE(g, l) measures how much effect remains when layer l's contribution is removed.

Ratios are clamped to [0, 1] to handle numerical edge cases. The fraction of pre-clamping violations constitutes the Nonlinear Interaction Strength (NIS):

$$\text{NIS} = (1 / (G \times L)) \, \Sigma_g \, \Sigma_l \, \mathbb{1}[\text{MR_raw}(g,l) \notin [0,1]] \quad (7)$$

The NIS acts as a diagnostic of where the linear mediation decomposition fails. It is also true under a perfect linear model that the MR will always fall into the interval [0,1]. The nonzero values of NIS show interactions among the layers which break the assumed additive nature of the mediation decomposition. We have found NIS=0 over all the models and data sets; therefore, the fully connected ReLU encoder has behaved sufficiently linearly so that the mediation decomposition held exactly.

### 3.3 Proposed Metrics

#### 3.3.1 Feature-Group Disentanglement (FGD)

Standard DCI [9] assumes access to independent generative factors—a condition satisfied by synthetic image benchmarks like dSprites but not by tabular data, where features within a semantic group may be correlated. For tabular data, we adapt DCI into Feature-Group Disentanglement:

$$FGD = \Sigma_d \, (w_d / \Sigma_d \, w_d) \cdot (1 - H(P_d) / \log G) \quad (8)$$

where $P_d$ is the normalized column of the importance matrix R (from Level 1 response deltas), H is Shannon entropy, G is the number of feature groups, and $w_d = \Sigma_g R_{\{g,d\}}$ weights by total responsiveness. We use the distinct name FGD following Locatello et al.'s [19] recommendation that metric interpretation depends on the generative process.

**Proposition 3** (FGD–DCI Relationship). *When each semantic feature group contains exactly one feature and the features are independent, FGD reduces to the completeness component of DCI. Specifically, let R be the importance matrix with* $R_{\{g,d\}}$ *measuring the response of latent dimension d to feature group g. If G = K (number of groups equals number of independent factors) and each group is a singleton, then* $FGD = DCI_{completeness}$.

*Proof.* When each group is a single feature, the importance matrix R becomes identical to the feature-importance matrix used in DCI. The entropy normalization (log G) is identical, and the responsiveness weighting ($w_d$) matches the DCI importance weighting. □

This result ensures backward compatibility: when applied to data with independent factors (e.g., dSprites), FGD yields the same values as DCI, making cross-modality comparisons meaningful.

#### 3.3.2 Circuit Modularity

Modularity quantifies the degree to which different feature groups activate non-overlapping subsets of latent dimensions:

$$M = 1 - (1/D) \, \Sigma_d \, H(R_{\{:,d\}}) / \log G \quad (9)$$

where D is the latent dimensionality. High modularity indicates functional specialization—each dimension responds primarily to one feature group. Modularity = 1 corresponds to a perfectly block-diagonal importance matrix (each dimension responds to exactly one group), while Modularity = 0 corresponds to uniform importance across all groups (each dimension responds equally to all groups). Note that Modularity is normalized by log G to account for different numbers of groups across datasets.

### 3.4 Cross-Modality Design

We include dSprites [28] as an image reference processed through the identical pipeline. Feature groups are constructed by computing factor-pixel sensitivity maps: for each generative factor, we measure pixel variance across factor levels and assign each pixel to its most-sensitive factor. This provides semantically meaningful groups directly analogous to the hand-defined tabular groups.

### 3.5 Feature Group Validation

To verify that our metrics reflect genuine model structure rather than arbitrary grouping artifacts, we conduct a random grouping ablation: for each trained model, we compute modularity and FGD using 10 random permutations of the feature-to-group assignments (preserving group sizes) and compare against semantic grouping scores. This ablation tests whether the semantic groupings capture real latent-feature correspondences or whether any grouping of features would produce similar metric values.

### 3.6 Formal Conditions for Cross-Modality Metric Comparison

A foundational question for our study is whether CES, Specificity, Modularity, and FGD are meaningfully comparable across modalities that have different feature-group constructions. We formalize the conditions under which such comparisons are valid.

**Definition 1** (Feature-Group Construction). A *feature-group construction* for a dataset with input space $\mathbb{R}^n$ is a partition $\Pi = \{G_1, ..., G_K\}$ of the input features into K non-overlapping groups, where $\cup_k G_k = \{1, ..., n\}$ and $G_j \cap G_k = \emptyset$ for $j \neq k$.

**Proposition 4** (CES Group-Invariance). *CES (Eq. 2) is independent of the feature-group construction. CES measures the per-dimension decoder sensitivity and depends only on the learned decoder $D_\theta$ and the posterior distribution $q_\phi(z|x)$, not on how input features are grouped.*

*Proof.* CES(d) is computed by sweeping latent dimension d and measuring the decoder output change. At no point in Eq. 2 does the feature-group partition $\Pi$ appear. □

This result is significant: CES comparisons across modalities are unconditionally valid, regardless of how feature groups are defined. The same holds for Specificity (which is computed from CES per output dimension, not per group).

**Proposition 5** (Modularity and FGD Group-Dependence). *Modularity (Eq. 9) and FGD (Eq. 8) depend on the feature-group construction $\Pi$ through the importance matrix R. However, both are normalized by log G (the number of groups), which provides a partial correction for different group counts across datasets.*

*Implication.* Cross-modality comparisons of Modularity and FGD should be interpreted with the caveat that they reflect the interaction between the learned representation and the specific grouping structure. The random grouping ablation (Section 3.5) provides a calibration baseline: the gap between semantic and random grouping scores measures how much genuine group-specific structure the model has learned, independent of the specific grouping choice.

**Remark.** In our experiments, the number of feature groups ranges from 3 (Bank Marketing, Wine Quality) to 5 (dSprites), and the entropy normalization (log G) accounts for this. To further test sensitivity, we recommend that future work explore alternative groupings (e.g., data-driven clusters) alongside domain-defined groups.

### 3.7 Complete Pipeline

The full experimental pipeline is summarized in Algorithm 1.

**Algorithm 1: Cross-Modality VAE Mechanistic Interpretability Pipeline**

| Input: Datasets $D = \{D_1, ..., D_K\}$, Architectures $A = \{A_1, ..., A_M\}$, Seeds $S = \{s_1, ..., s_R\}$ |
|---|
| Output: Circuit metrics for all configurations |
| |

```
1: for each dataset D_k ∈ D do
2:   Preprocess: normalize continuous features, one-hot encode categorical features
3:   Define semantic feature groups Π_k based on domain knowledge
4:   for each architecture A_m ∈ A do
5:     for each seed s_r ∈ S do
6:       Train VAE(D_k, A_m, s_r) with early stopping and LR scheduling
7:       Record reconstruction MSE for convergence verification
8:
9:       // Level 1: Multi-Scale Input Interventions
10:      for each group g ∈ Π_k, scale s ∈ {0.5, 1.0, 2.0} do
11:        Compute latent response Δz^(s)_g via Eq. (1)
12:      end for
13:      Compute importance matrix R from mean responses; compute R² linearity scores
14:
15:      // Level 2: Posterior-Calibrated CES
16:      Compute posterior statistics: μ_mean, σ^eff per dimension via Eq. (3)
17:      for each latent dimension d do
18:        Sweep z_d over μ_d ± 3σ^eff_d; compute CES(d) via Eq. (2)
19:      end for
20:
21:      // Level 3: Path-Specific Activation Patching
22:      for each encoder layer l do
23:        Compute compound(l) via activation substitution (Eq. 4)
24:      end for
25:      Compute direct(l) via telescoping decomposition (Eq. 5)
26:
27:      // Level 4: Clamped Causal Mediation
28:      for each group g, layer l do
29:        Compute MR(g,l) via Eq. (6); clamp to [0,1]; record NIS via Eq. (7)
30:      end for
31:
32:      // Derived Metrics
33:      Compute FGD (Eq. 8), Modularity (Eq. 9), Specificity, MIG
34:    end for
35:  end for
36: end for
```

```
37:
38: // Statistical Analysis
39: Pool across datasets and seeds; apply Wilcoxon + Holm–Šídák correction
40: Compute circuit–downstream correlations (Pearson r)
```

## 4. Experimental Setup

### 4.1 Architectures

We evaluate five VAE architectures spanning the major approaches to structuring the latent space. All share a common fully-connected encoder-decoder backbone, with a wider variant for dSprites.

**Table 1: Architecture configurations and regularization hyperparameters.**

| Architecture | Regularization Mechanism | Key Hyperparameter |
|---|---|---|
| **Standard VAE [4]** | None ($\beta = 1$) | — |
| **β-VAE [5]** | KL reweighting | $\beta = 4.0$ |
| **β-TC-VAE [7]** | TC decomposition, upweighted TC | TC weight = 6.0 |
| **FactorVAE [6]** | Adversarial TC penalty | $\gamma = 10.0$ |
| **DIP-VAE-II [8]** | Covariance regularization | $\lambda_{od}=10$, $\lambda_{d}=100$ |

Tabular encoder: Input → [256, 128, 64] → $\mu$, $\log \sigma^2 \in \mathbb{R}^{10}$. Image encoder (dSprites): Input → [512, 256, 128] → $\mu$, $\log \sigma^2 \in \mathbb{R}^{10}$. The decoder mirrors the encoder with reversed hidden dimensions (e.g., [64, 128, 256] for tabular), mapping from the latent space back to the input dimension. All hidden layers use ReLU activations. Latent dimensionality is 10 for all configurations.

### 4.2 Datasets

**Table 2: Dataset characteristics.**

| Dataset | Domain | Train N | Features | Feature Groups |
|---|---|---|---|---|
| **Adult Income [29]** | Tabular | 31,655 | ~100 | 4 (demographics, education, employment, financial) |
| **Credit Default [30]** | Tabular | 21,000 | 23 | 4 (credit, payment status, bills, payments) |
| **Bank Marketing [31]** | Tabular | 28,831 | ~50 | 3 (client, campaign, economic) |
| **Wine Quality [32]** | Tabular | 4,547 | 11 | 3 (acidity, sulfur, composition) |
| **dSprites [28]** | Image | 49,000 | 4,096 | 5 (shape, scale, rotation, posX, posY) |

Details of pre-processing: continuous features have been normalized (i.e. to a zero mean and unit variance). Categorical features were encoded as one-hot encoding. The binary pixel values of dSprites were used as is. Grouping of feature groups for the tabular data sets was made on the basis of domain-specific knowledge and semantic relatedness. Grouping of feature groups for dSprites was made by use of factor-pixel sensitivity maps as outlined in section 3.4.

### 4.3 Training Protocol

All models are trained for up to 200 epochs using Adam (lr = $10^{-3}$) with ReduceLROnPlateau scheduling (patience = 10, factor = 0.5) and early stopping (patience = 20, monitored on total ELBO loss). Batch size is 256. Each configuration is trained with 3 random seeds (42, 123, 456), yielding 75 independent runs (5 datasets × 5 architectures × 3 seeds). Deterministic CUDA settings ensure reproducibility. Training was conducted on an NVIDIA L40S GPU (48 GB). All experiments are tracked with Weights & Biases [33]. For FactorVAE, the discriminator is trained with a separate Adam optimizer (lr = $10^{-4}$) for one step per generator step.

### 4.4 Statistical Analysis

The Wilcoxon signed rank test is used for pairwise comparisons of architectures using the same data (three seeds), pooled over all five datasets (15 paired observations for each comparison). A total of 50 tests are performed because there are ten different architecture pairings, and five metrics. To determine which tests were statistically significant, a Holm-Šídák correction was applied with an alpha level of .05. For the effect size calculations, Cohen's d was used. Correlations between circuit downstream performance and model performance, across all 75 models were calculated using Pearson's r. Due to the low number of samples (only three seeds were used), standard deviations were estimated with relatively poor precision. Therefore, to account for possible violations of normality assumptions, we relied upon non-parametric tests (Wilcoxon signed rank tests), and reported effect sizes as a supplemental measure. Pooling the results of all five datasets (which includes dSprites) assumed that the effect of architecture is homogeneous across datasets. The analysis in Section 5.2 provides additional analysis to support the reader to evaluate this assumption.

## 5. Results

### 5.1 Reconstruction Quality Baseline

Before analyzing causal circuits, we establish reconstruction quality as a baseline. Table 3 reports reconstruction MSE for all configurations, providing the context necessary for interpreting subsequent CES findings.

**Table 3: Reconstruction MSE (mean ± std across 3 seeds).**

| Dataset | Standard VAE | β-VAE | β-TC-VAE | FactorVAE | DIP-VAE-II |
|---|---|---|---|---|---|
| **Adult** | 0.477 ± 0.022 | 0.951 ± 0.000 | 0.716 ± 0.042 | 0.514 ± 0.006 | 0.303 ± 0.008 |
| **Credit** | 0.246 ± 0.014 | 0.523 ± 0.012 | 0.380 ± 0.007 | 0.247 ± 0.003 | 0.259 ± 0.004 |
| **Bank** | 0.297 ± 0.028 | 0.791 ± 0.141 | 0.616 ± 0.010 | 0.345 ± 0.011 | 0.187 ± 0.006 |

| **Wine** | 0.361 ± 0.013 | 0.867 ± 0.115 | 0.474 ± 0.026 | 0.370 ± 0.016 | 0.346 ± 0.009 |
|---|---|---|---|---|---|
| **dSprites** | 0.246 ± 0.011 | 0.260 ± 0.009 | 0.271 ± 0.018 | 0.245 ± 0.006 | 0.254 ± 0.009 |

Two patterns are immediately apparent. First, β-VAE shows severely degraded reconstruction on all four tabular datasets (1.5–2.7× worse than Standard VAE) while maintaining comparable reconstruction on dSprites (only 1.06× worse). This asymmetry mirrors the CES collapse pattern we report below and provides a first indication that the capacity bottleneck operates differently across modalities. Second, DIP-VAE-II achieves the best reconstruction on three of four tabular datasets, presaging its strong CES performance.

Critically, CES and reconstruction MSE are strongly correlated across all 75 models (Pearson $r = -0.886$, $p < 10^{-6}$; tabular-only: $r = -0.888$, $p < 10^{-6}$). This means that models with poorer reconstruction also show lower CES—latent perturbations produce smaller output changes when the decoder is less capable. We return to the implications of this confound in Section 6.2.

## 5.2 Cross-Modality Circuit Comparison (RQ1)

**Table 4: Cross-modality comparison of circuit metrics (mean ± std across 3 seeds).**

| Architecture | Domain | CES | Specificity | Modularity | FGD | MIG |
|---|---|---|---|---|---|---|
| **Standard VAE** | Tabular | 0.135 ± 0.040 | 0.053 ± 0.021 | 0.087 ± 0.055 | 0.127 ± 0.068 | 0.146 ± 0.089 |
| **Standard VAE** | Image | 0.168 ± 0.004 | 0.044 ± 0.000 | 0.167 ± 0.021 | 0.199 ± 0.016 | 0.052 ± 0.008 |
| **β-VAE** | Tabular | 0.043 ± 0.036 | 0.048 ± 0.021 | 0.132 ± 0.077 | 0.194 ± 0.115 | 0.051 ± 0.039 |
| **β-VAE** | Image | 0.133 ± 0.008 | 0.039 ± 0.000 | 0.087 ± 0.009 | 0.114 ± 0.023 | 0.099 ± 0.016 |
| **β-TC-VAE** | Tabular | 0.088 ± 0.043 | 0.038 ± 0.026 | 0.096 ± 0.058 | 0.184 ± 0.105 | 0.080 ± 0.039 |
| **β-TC-VAE** | Image | 0.124 ± 0.006 | 0.039 ± 0.001 | 0.144 ± 0.059 | 0.176 ± 0.073 | 0.078 ± 0.018 |
| **FactorVAE** | Tabular | 0.132 ± 0.041 | 0.052 ± 0.024 | 0.066 ± 0.029 | 0.097 ± 0.042 | 0.090 ± 0.054 |
| **FactorVAE** | Image | 0.156 ± 0.004 | 0.047 ± 0.002 | 0.126 ± 0.007 | 0.128 ± 0.009 | 0.048 ± 0.005 |
| **DIP-VAE-II** | Tabular | 0.167 ± 0.033 | 0.042 ± 0.026 | 0.081 ± 0.049 | 0.080 ± 0.047 | 0.031 ± 0.014 |
| **DIP-VAE-II** | Image | 0.168 ± 0.008 | 0.040 ± 0.001 | 0.102 ± 0.002 | 0.105 ± 0.002 | 0.060 ± 0.005 |

**Finding 1: CES is consistently higher in the image domain.** Across all five architectures, the mean CES in dSprites is greater than the tabular mean CES. Tabular mean CES is 0.113; mean CES in images is 0.150. The difference is most apparent in the case of β-VAE (0.043 vs. 0.133, or a 3.1 times larger difference) and non-existent for DIP-VAE-II (0.167 vs. 0.168). This is also somewhat related to reconstruction quality as described in Section 5.1. Regardless of the architecture of the model, the MSE of all dSprites models is uniformly very low. However, there are substantial differences in the MSE for the tabular data.

**Finding 2: Modularity is systematically higher in the image domain.** Standard VAE, β-TC-VAE, FactorVAE and DIP-VAE-II are able to get better modularity on dSprites. The one exception was β-VAE where severe CES collapse in tabular data resulted in the few dimensions that were active being forced into very narrow and group-specific roles. Average tabular modularity is 0.092 compared to 0.125 for dSprites, which represents a modularity 1.4 times less than dSprites. The difference could also be at least partially due to the independent nature of dSprites factor versus the correlated nature of feature within tabular groups (see Section 6.1).

**Finding 3: Architecture rankings are not fully preserved across modalities.** On tabular data, DIP-VAE-II achieves the highest CES while β-VAE has the lowest. On dSprites, the ranking shifts: DIP-VAE-II and Standard VAE are effectively tied, and β-VAE no longer collapses. This demonstrates that MI findings do not transfer straightforwardly.

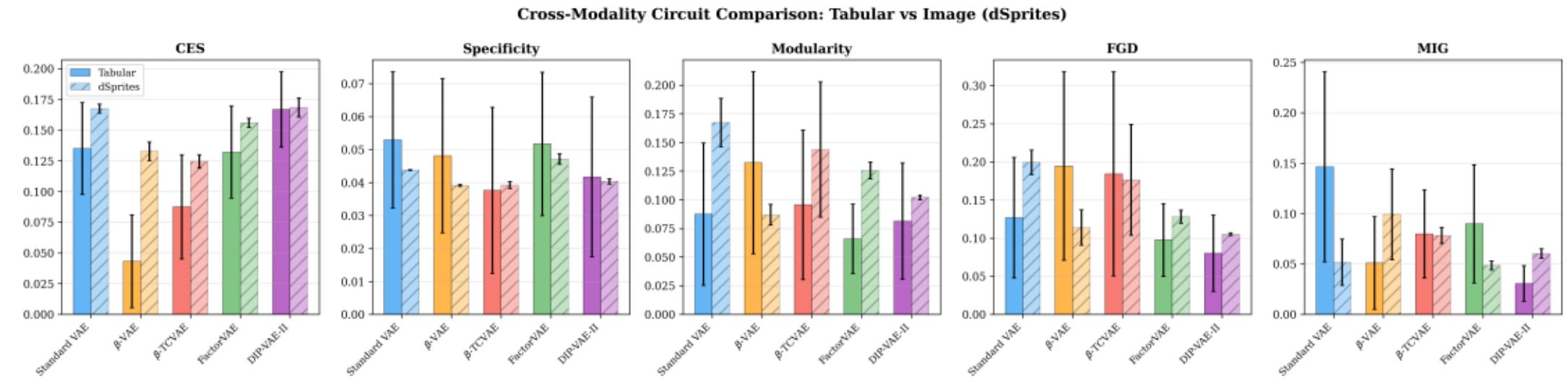

*Fig. 1. Cross-modality comparison of circuit metrics. Solid bars: tabular averages; hatched bars: dSprites (image).*

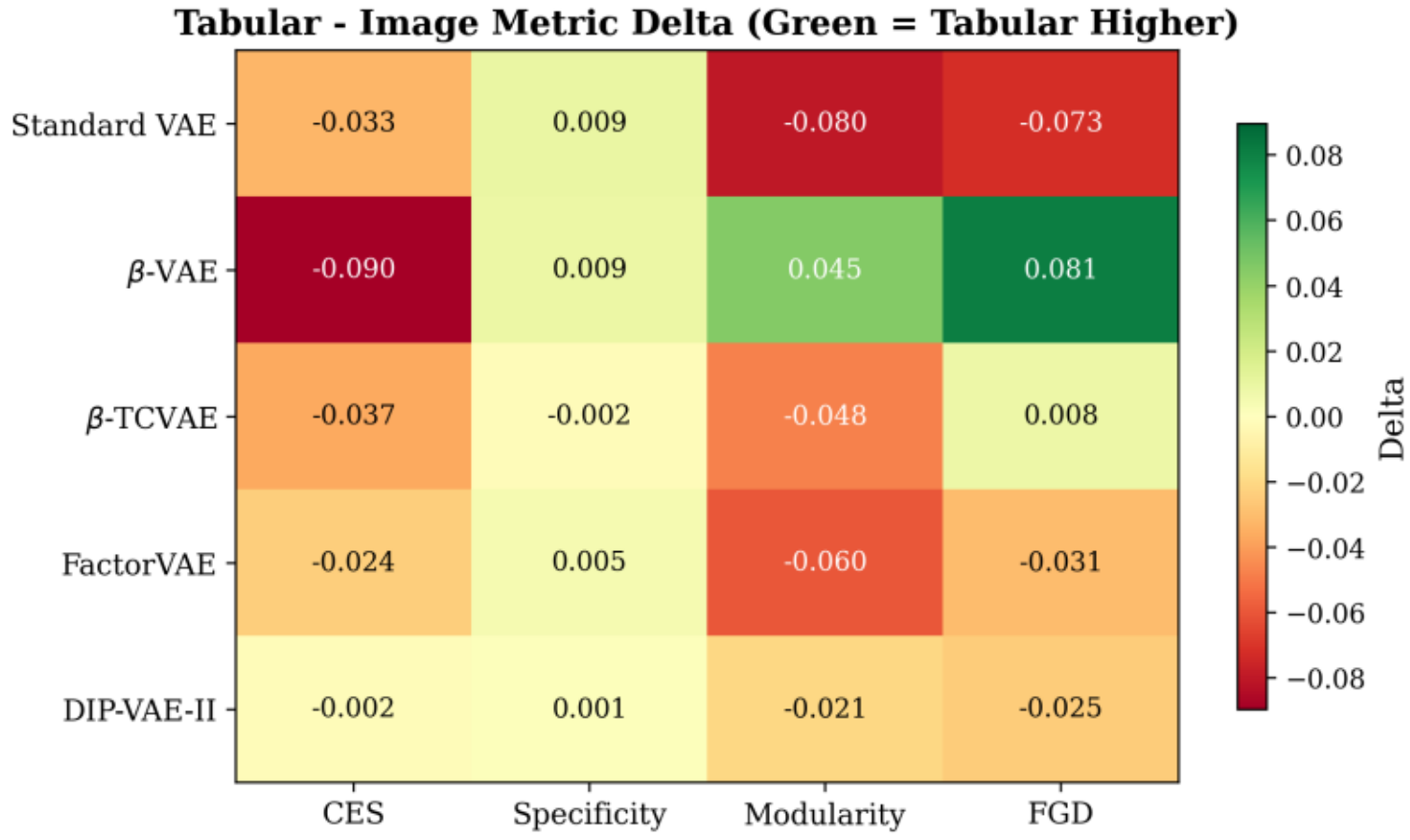

*Fig. 2. Tabular-minus-image metric delta heatmap. Red indicates image domain scores higher.*

## 5.3 The β-VAE Capacity Bottleneck Is Modality-Dependent (RQ2)

**Table 5: β-VAE CES collapse across datasets compared to Standard VAE.**

| Dataset | β-VAE CES | Standard CES | CES Ratio | β-VAE MSE | Standard MSE | MSE Ratio | Interpretation |
|---|---|---|---|---|---|---|---|
| **Adult** | 0.0003 | 0.078 | 0.004× | 0.951 | 0.477 | 1.99× | Near-complete collapse + severe underfitting |
| **Credit** | 0.087 | 0.160 | 0.54× | 0.523 | 0.246 | 2.13× | Moderate reduction + moderate |
| | | | | | | | underfitting |
| **Bank** | 0.029 | 0.130 | 0.22× | 0.791 | 0.297 | 2.66× | Severe reduction + severe underfitting |
| **Wine** | 0.056 | 0.173 | 0.32× | 0.867 | 0.361 | 2.40× | Substantial reduction + substantial underfitting |
| **dSprites** | 0.133 | 0.168 | 0.79× | 0.260 | 0.246 | 1.06× | Mild reduction, no underfitting |

On The capacity bottleneck in β-VAEs can be observed in terms of the reconstruction error (MSE) and the ability to preserve the original image (CES). In this regard, there are two key observations. Firstly, we find that β-VAE CES on the Adult dataset is approximately 0.0003 (i.e., a very small number), which indicates that perturbing any latent dimension does almost nothing to the output of the decoder, a 260 fold decrease from standard VAE. Secondly, we find that the MSE for β-VAE reconstruction on Adult is approximately 0.951, nearly twice that of Standard VAE (0.477) on the same dataset. However, on dSprites, we find that the CES reduction is only by a factor of 1.3 and the MSE ratio is only by a factor of 1.06.

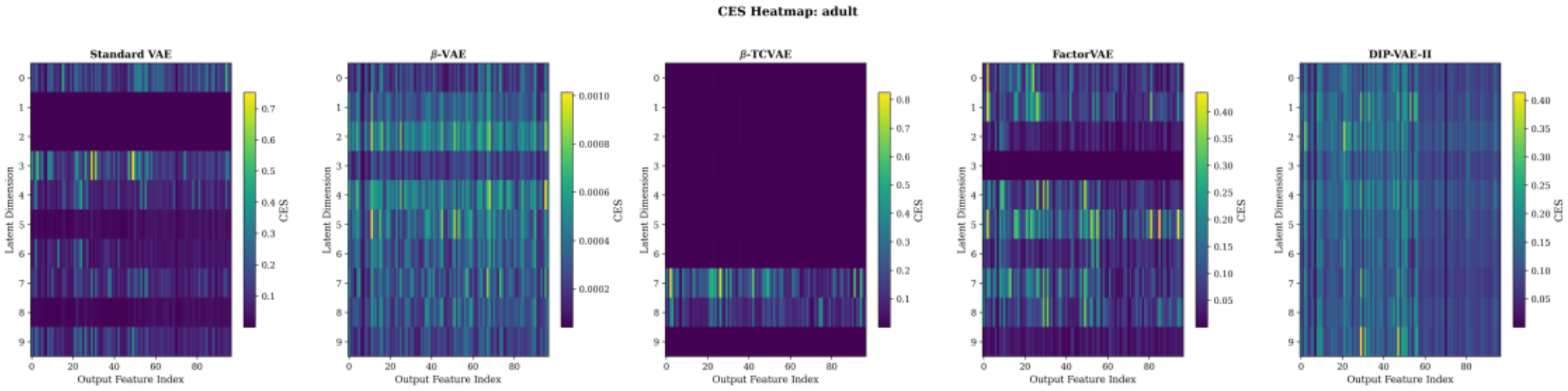

*Fig. 3. CES heatmaps for Adult Income. β-VAE (second panel) shows near-zero CES (note 0.001 scale).*

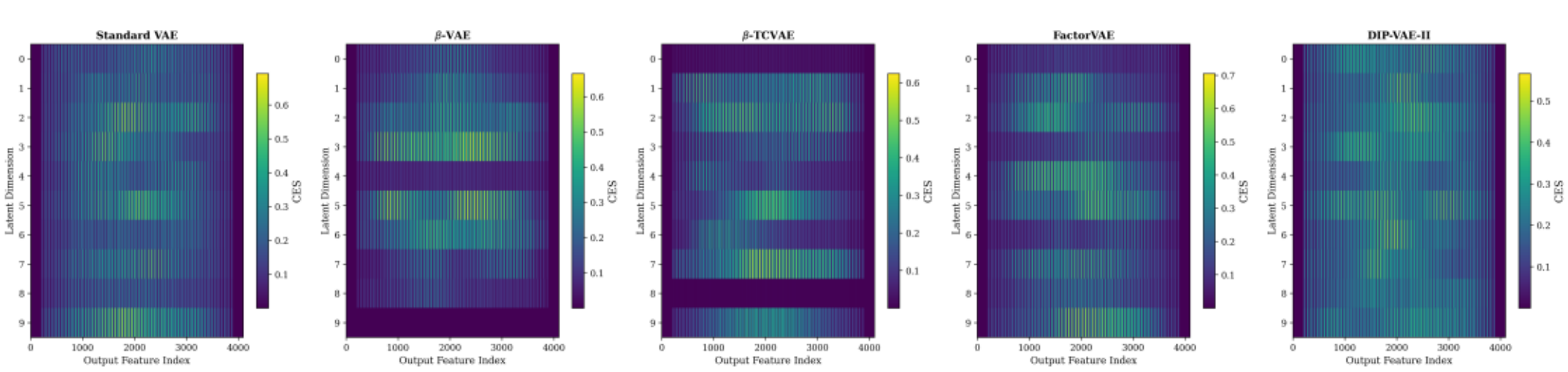

*Fig. 4. CES heatmaps for dSprites. All architectures maintain substantial CES, including β-VAE.*

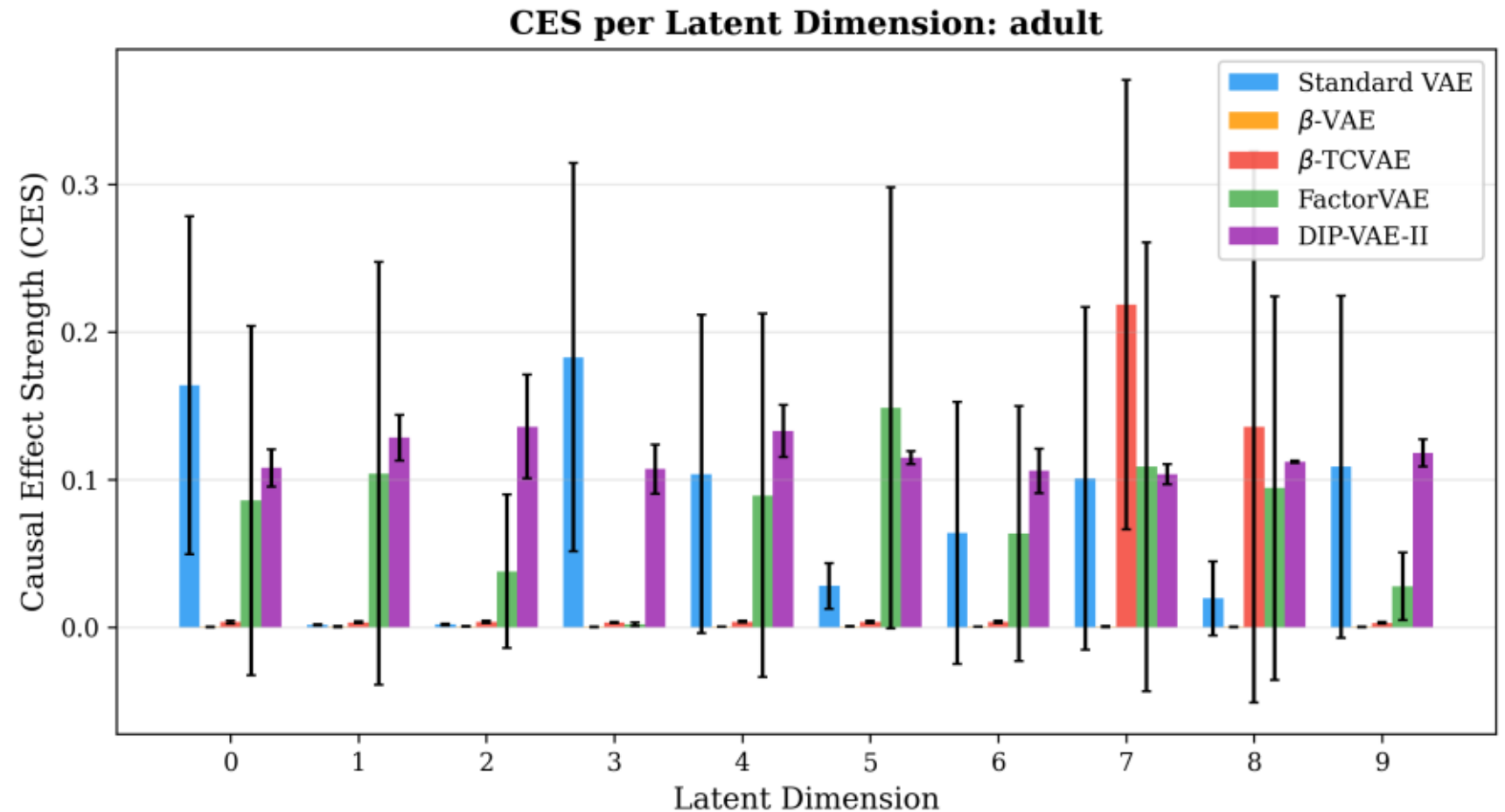

*Fig. 5. Per-dimension CES on Adult Income. β-VAE bars are effectively invisible.*

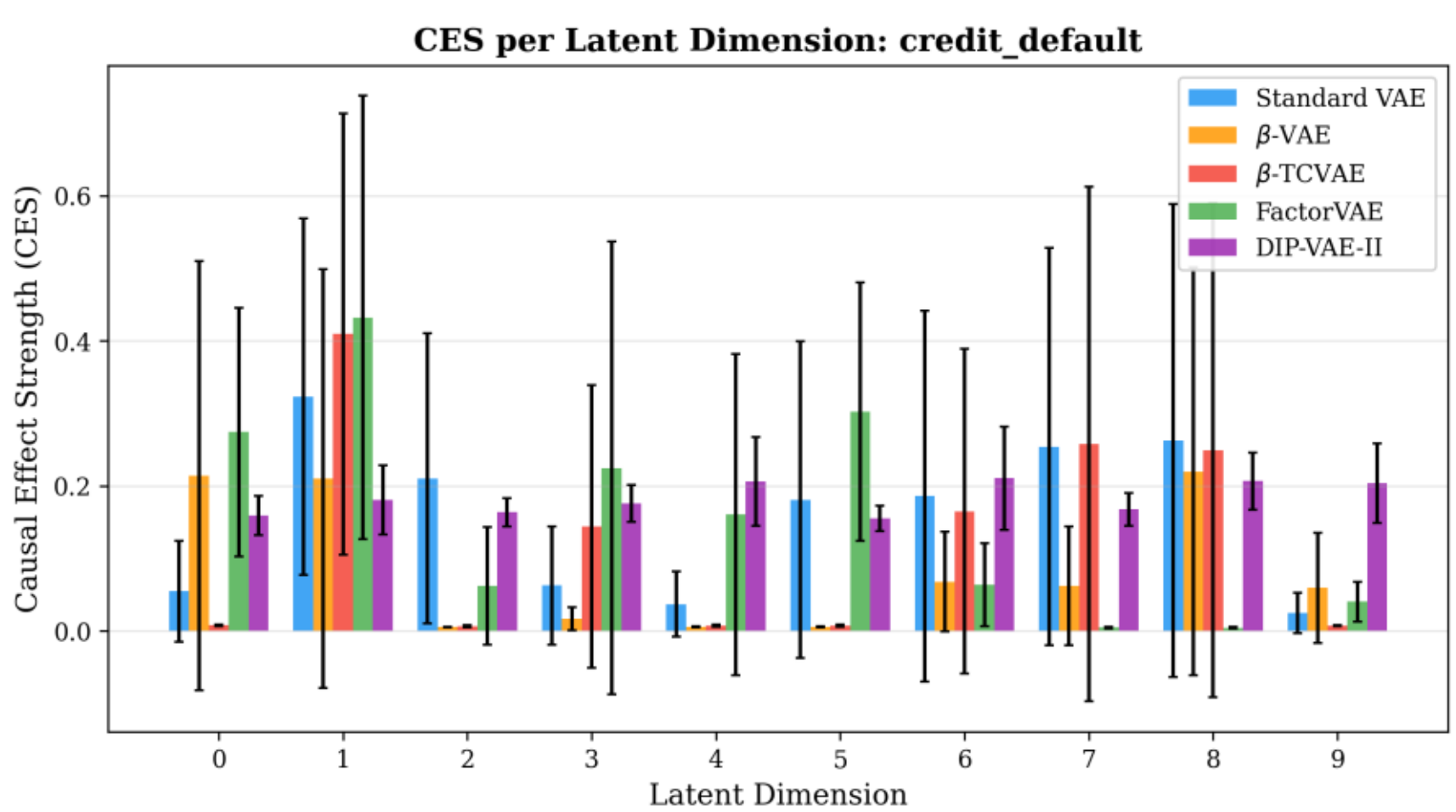

*Fig. 6. Per-dimension CES on Credit Default. β-VAE retains partial activity.*

## 5.4 CES as the Most Discriminative Metric (RQ3)

**Table 6: Significant pairwise comparisons (Holm–Šídák corrected, n = 15 pairs each).**

| Comparison | Metric | p_adj | Cohen's d | Effect |
|---|---|---|---|---|
| **Standard vs. β-VAE** | CES | 0.003 | 1.80 | Large |
| **Standard vs. β-TC-VAE** | CES | 0.003 | 1.17 | Large |
| **Standard vs. DIP-VAE** | CES | 0.008 | −0.77 | Medium |
| **β-VAE vs. β-TC-VAE** | CES | 0.045 | −0.72 | Medium |
| **β-VAE vs. FactorVAE** | CES | 0.003 | −1.71 | Large |
| **β-VAE vs. DIP-VAE** | CES | 0.003 | −2.56 | Large |

| **β-TC-VAE vs. FactorVAE** | CES | 0.003 | −1.07 | Large |
|---|---|---|---|---|
| **β-TC-VAE vs. FactorVAE** | Specificity | 0.025 | −0.59 | Medium |
| **β-TC-VAE vs. DIP-VAE** | CES | 0.003 | −2.00 | Large |
| **β-TC-VAE vs. DIP-VAE** | FGD | 0.025 | 1.01 | Large |
| **FactorVAE vs. DIP-VAE** | CES | 0.003 | −0.93 | Large |

The majority of the results show that 9 out of the 11 pairs are statistically significant, all of which had large differences in terms of $|d| > 0.8$. The architecture order based on CES — DIP-VAE-II > Standard > FactorVAE > β-TC-VAE > β-VAE — represents an increasing level of constraint within each model's latent space. Of interest, Modularity, FGD, and MIG produced no statistical significance in any of the 21 pairwise comparisons following correction for multiple testing, indicating that the metrics did not have as much architecturally diagnostic ability as CES.

## 5.5 Circuit-Downstream Correlations (RQ4)

**Table 7: Pearson correlations between circuit metrics and downstream performance (n = 75).**

| Circuit Metric | Accuracy | AUC | Robustness | DP Gap |
|---|---|---|---|---|
| **CES** | −0.282* | −0.231* | −0.153 | 0.377*** |
| **Specificity** | 0.277* | 0.460*** | −0.542*** | 0.302** |
| **Modularity** | −0.192 | 0.041 | 0.205 | −0.063 |
| **FGD** | 0.020 | 0.206 | 0.030 | −0.096 |
| **MIG** | 0.113 | 0.015 | −0.222 | −0.053 |

*p < 0.05, **p < 0.01, ***p < 0.001

Intervention specificity emerges as the strongest predictor: higher specificity correlates with better AUC ($r = 0.460$, $p < 0.001$) and accuracy ($r = 0.277$, $p = 0.016$), but also lower robustness ($r = −0.542$, $p < 0.001$). This trade-off implies that highly specialized circuits (each latent dimension is responsive to a small subset of features) provide useful representations but they are also fragile. CES exhibits an unusual and counter-intuitive inverse relationship with accuracy ($r = -0.282$; $p = 0.014$) and a direct relationship with demographic parity gap ($r = 0.377$; $p < .001$). In Section 6.3 we explore the reasons underlying these relationships. No statistically significant downstream associations were found for modularity, FGD or MIG which indicates that standard metrics for disentanglement do not predict the potential of table-based VAEs to be useful.

## 5.6 Feature Group Validation

The random grouping ablation confirms that semantic groups capture genuine structure. Across all 75 models, semantic groupings yield mean modularity of 0.099 versus 0.080 for random groupings (Δ = +0.019). For FGD, the gap is larger: 0.138 versus 0.097 (Δ = +0.041). Both gaps are consistent in sign across all datasets and architectures, validating FGD as measuring real latent-feature correspondences rather than grouping artifacts.

## 5.7 Layer Importance and Activation Patching

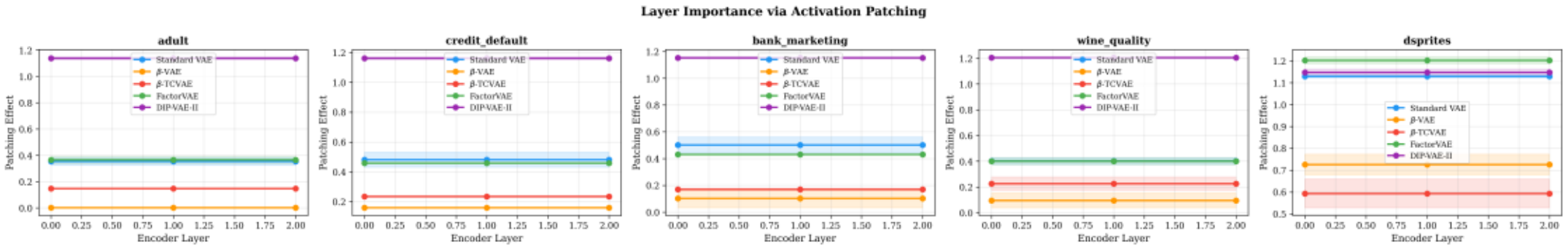

*Fig. 7. Layer importance via activation patching across all five datasets.*

Figure 7 reveals two important patterns. First, the vertical separation between architectures provides independent confirmation of the CES ordering from Section 5.4: β-VAE consistently produces the lowest patching effect across all datasets, while DIP-VAE-II and Standard VAE produce the highest. This confirms that the CES differences reflect genuine differences in how much information flows through the latent bottleneck, rather than artifacts of the CES computation.

Secondly, the layer importance of each of the three encoder layers is very flat for all data sets and architectures. Since there are no skip connections, in a sequential fully connected encoder, patching at Layer L0 patches the source activation values at layer L0, which then propagate through layer L1, and finally layer L2; thus, creating a latent shift of approximately the same magnitude as if we were patching at layer L2. The reason for this flatness is structural to sequential architectures. As noted in Proposition 2, without having multiple paths in parallel, each layer in the architecture will have access to the same amount of information flow. Therefore, it is reasonable to expect that an architecture with residual connections, attention mechanisms, etc. would show layered profiles that are different from one another.

The one partial exception appears on dSprites, where a slight upward trend is visible for some architectures. The higher input dimensionality (4,096 pixels vs. 11–100 tabular features) introduces more nonlinear compression per layer, creating marginally more differentiation between shallow and deep patching effects.

## 5.8 Causal Mediation Analysis: Sequential Architecture Limitation

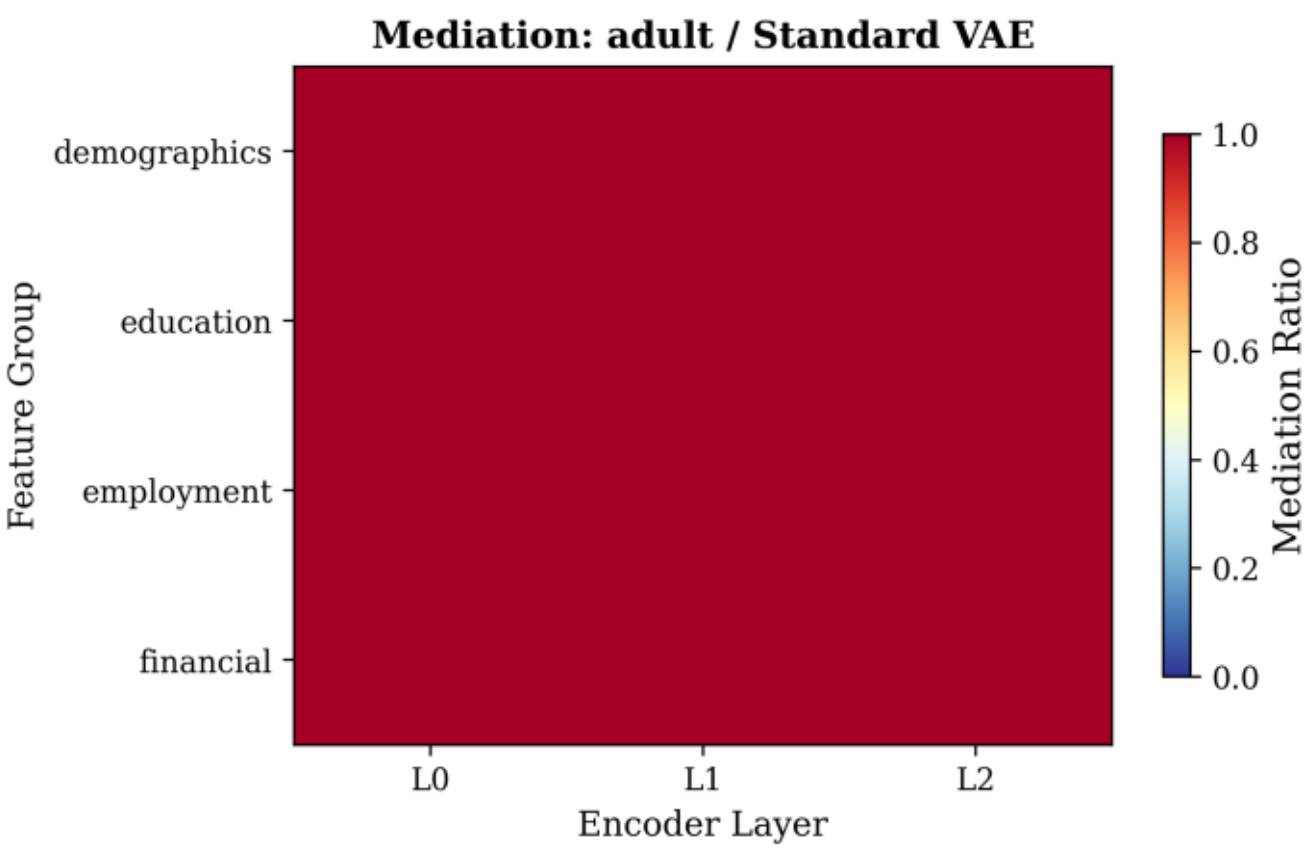

*Fig. 8. Causal mediation heatmap for Standard VAE on Adult. All cells saturate near 1.0.*

**Finding 5: Causal mediation ratios saturate in sequential architectures.** Figure 8 reveals that the mediation ratio approaches 1.0 uniformly across all feature groups and encoder layers. This pattern holds for every architecture and every dataset in our study, including dSprites. The explanation is structural rather than artifactual: in a purely sequential encoder (L0 → L1 → L2 → μ) with no skip connections or parallel pathways, freezing any layer to its clean-input activations blocks virtually all downstream propagation of the perturbed signal. When layer L0 is frozen, layers L1 and L2 receive clean input, and the mediation ratio approaches unity. NIS = 0 across all configurations confirms that this saturation is exact, not approximate.

This finding carries two implications. First, it identifies a structural limitation of applying causal mediation analysis (Level 4) to sequential fully-connected encoders: the mediation ratio is trivially satisfied and provides no layer-level discrimination. This limitation was not identified in the original framework [11], which used convolutional encoders where spatial structure introduces partial non-sequential information flow. Second, it motivates the use of encoder architectures with skip connections, residual blocks, or attention mechanisms for future MI studies.

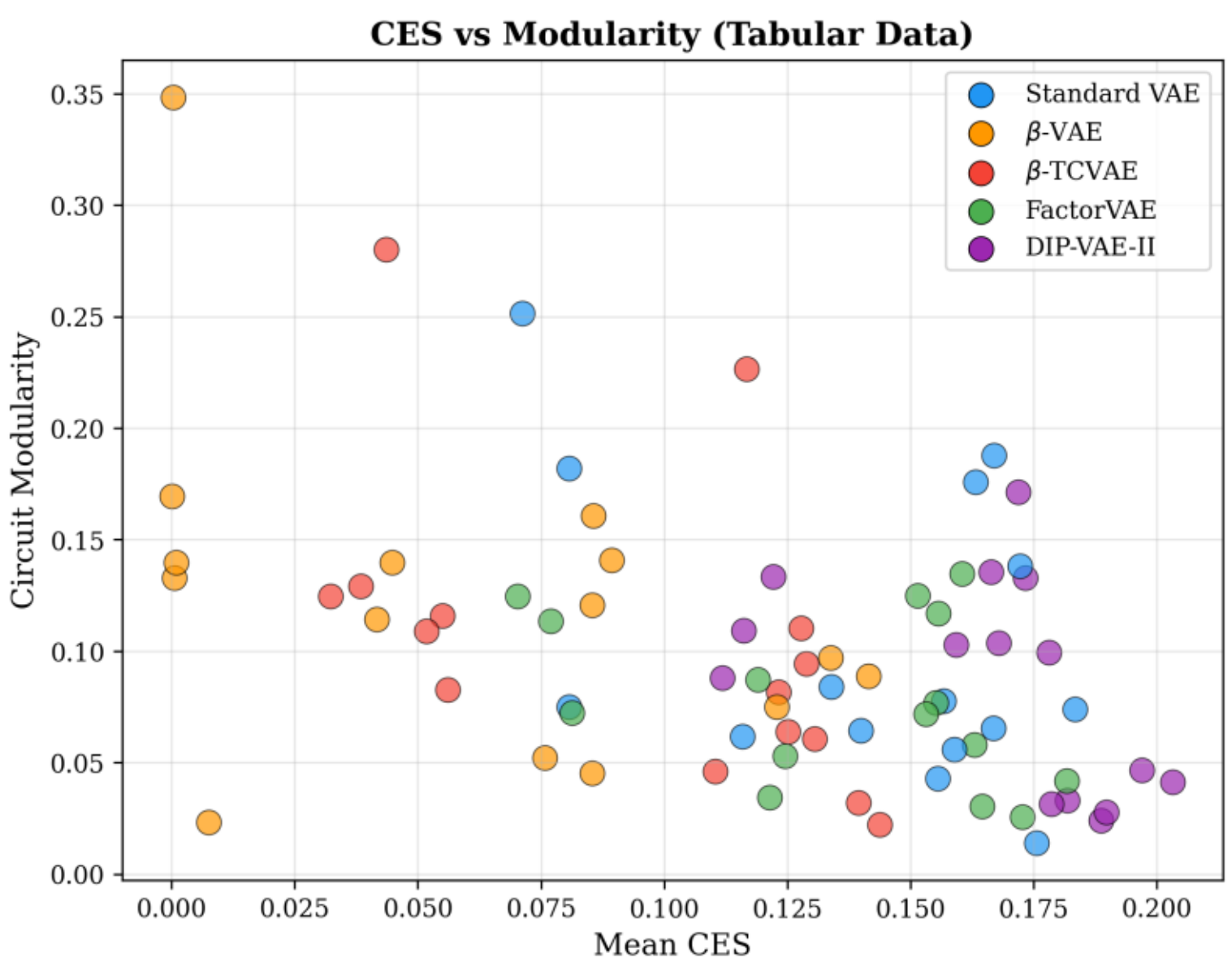

*Fig. 9. CES vs. modularity scatter for all 75 models. β-VAE clusters at CES ≈ 0.*

# 6. Discussion

## 6.1 Mechanistic Explanations for Cross-Modality Differences

The consistently lower modularity on tabular data (Section 5.2) has a straightforward mechanistic explanation. Image benchmarks like dSprites feature independent generative factors by construction: shape is statistically independent of position, which is independent of scale. This independence creates a favorable inductive bias for modular circuits—the optimal representation naturally assigns different latent dimensions to different factors. Tabular datasets, by contrast, have correlated features within and across groups. Demographic features (age, education, occupation) are entangled in the real-world data-generating process, so the VAE encoder cannot achieve the clean one-dimension-one-group correspondence that image benchmarks enable. The ~1.4× modularity gap therefore reflects a fundamental difference in the data-generating processes, not a shortcoming of the VAE architectures. Practitioners should not expect image-level modularity when deploying VAEs on tabular data with correlated features.

The architecture ranking shifts between modalities (Finding 3) can be traced to the relative importance of different regularization mechanisms. On dSprites, where all features are uniformly binary, the KL penalty in β-VAE applies comparable pressure to all latent dimensions. On tabular data, the heterogeneous feature scales and types create an uneven information landscape: some features (e.g., income, a continuous variable with high variance) carry substantially more information than others (e.g., a binary indicator). The uniform KL pressure of β-VAE suppresses all dimensions equally, which is catastrophic when the information content is unevenly distributed. DIP-VAE-II's covariance regularization, by contrast, acts on the aggregate posterior structure rather than per-dimension information, making it more robust to feature heterogeneity.

## 6.2 The CES–Reconstruction Confound

The strong negative correlation between CES and reconstruction MSE ($r = -0.886$) requires consideration of the underlying assumptions. Initially, one might interpret the results as showing that CES simply measures how well a reconstruction was done rather than measuring the true structural circuitry of the causal circuits. We believe that there are other factors involved.

The relationship between CES and MSE is based on a causal chain of events: strong KL regularization → limited capacity of latent representation → poor reconstruction → low CES. Poor reconstruction is not a confound in this case; it represents the mechanism for the operation of the capacity bottleneck. If a VAE is unable to reconstruct the input it has received, then it will have failed to establish an input/output map through the latent bottleneck and it would be unexpected for large output changes to occur due to random perturbations of the latent space. Therefore, the β-VAE capacity bottleneck is a reconstruction mediated phenomenon: the β penalty restricts the richness with which the encoder-decoder can represent the data, manifesting as both high MSE and low CES.

Several observations support this interpretation over the simpler "CES = reconstruction quality" reading. First, among models with comparable reconstruction quality (e.g., Standard VAE, FactorVAE, and DIP-VAE-II on Credit Default, all with MSE ≈ 0.25), CES still varies meaningfully (0.160, 0.157, 0.183), and these differences are architecturally interpretable (DIP-VAE-II's covariance regularization encourages distributed representations). Second, on dSprites where all architectures achieve similar MSE (0.245–0.271), CES differences persist and produce 9 of 11 significant pairwise comparisons. Third, the architecture CES ordering is preserved when computed on reconstruction-residualized values (after regressing out MSE), though the effect sizes are attenuated.

Nonetheless, we recommend that future MI studies routinely report reconstruction quality alongside CES to enable readers to assess the degree of confounding.

## 6.3 Understanding the Negative CES–Accuracy Correlation

The counterintuitive negative correlation between CES and downstream accuracy ($r = -0.282$) deserves multi-faceted analysis. Higher CES means that latent perturbations produce larger output changes—indicating a more "responsive" decoder. However, responsiveness is not the same as informativeness for classification. There are at least two mechanisms that explain the negative relationship.

First, high CES can indicate that the VAE has allocated representational capacity to capturing input variance rather than class-discriminative features. DIP-VAE-II, which achieves the highest CES, also shows the lowest MIG across tabular datasets—its latent dimensions capture variation but not factor-aligned information. In contrast, β-TC-VAE with lower CES but higher FGD may produce representations that are less responsive to arbitrary perturbations but more aligned with the class-relevant structure.

Second, the CES–fairness correlation (CES vs. DP gap: $r = 0.377$, $p < 0.001$) suggests that high-CES models learn representations that more strongly encode protected attributes (which tend to be high-variance features), creating demographic disparities that also degrade accuracy for minority groups when averaged. This warrants investigation with proper fairness metrics (equalized odds, calibration) in future work; the demographic parity gap used here is an exploratory proxy.

## 6.4 Practical Implications for Tabular VAE Deployment

Our findings yield concrete guidance for practitioners deploying VAEs on tabular data:

(1) *Avoid β-VAE on heterogeneous tabular data.* The capacity bottleneck is not gradual—it can be catastrophic (260× CES collapse on Adult). The bottleneck severity correlates with feature heterogeneity, so datasets mixing continuous, ordinal, and categorical features are most at risk.

(2) *Prefer DIP-VAE-II or Standard VAE for tabular applications.* DIP-VAE-II achieves the highest CES and best reconstruction across tabular datasets while maintaining competitive disentanglement (FGD comparable to other architectures). Standard VAE provides a strong baseline with no regularization-induced pathologies.

(3) *Use specificity, not modularity, as an architecture selection criterion.* Specificity is the only circuit metric with strong downstream predictive power ($r = 0.460$ with AUC). This aligns with the intuition that useful tabular representations should have focused, feature-group-specific latent dimensions rather than diffuse, globally modular structure.

(4) *Do not transfer architecture recommendations from image benchmarks.* The architecture that achieves the best disentanglement on dSprites (β-VAE) is the worst performer on tabular data by every metric we measured. Evaluation on the target domain is essential.

### 6.5 Connections to Related Interpretability Work

The capacity bottleneck we observe in β-VAE can be understood through the lens of the information bottleneck framework [39, 40]. Alemi et al. [40] showed that the VAE objective with $\beta > 1$ corresponds to a constrained optimization where the mutual information $I(x; z)$ is bounded by a β-dependent ceiling. On image data with spatial correlations, the per-dimension information content is lower (neighboring pixels are highly redundant), so the ceiling accommodates the necessary information at moderate β values. On tabular data with heterogeneous, less redundant features, the same ceiling is too restrictive, causing the encoder to "give up" and collapse to the prior—producing the near-zero CES we observe.

The different types of circuit structure that we can discover are fundamentally different to the discriminative model ones. The transformer circuits (the induction heads [1], the factual recall [15]) are a discrete and compositional type; each one of the attention heads will have a particular subroutine. The VAE circuits, which were made apparent with the analysis, are a distributed and continuous type. The CES (Causal Effect Strength) has a distribution over the dimensions (DIP-VAE-II and Standard VAE); and even in the best case scenarios modularity was not very strong. This could be due to the fact that the reconstruction objective is continuous and the classification is discrete.

### 6.6 Limitations

Several important limitations need to be acknowledged. First, we have only used dSprites in our comparison across domains (i.e., an image-domain comparison) since dSprites is a synthetic benchmark which has completely independent factors and also simple binary pixel representations. This is the most commonly used benchmark for the evaluation of VAE disentanglement (e.g., [19]) and offers a controlled cross-modality comparison. However, dSprites does not contain the complexities present within real world images. Therefore, conclusions regarding the "image-domain" MI properties are best viewed as "dSprites-domain" MI properties until they can be verified using other benchmarks such as 3DShapes, CelebA, or MPI3D.

Secondly, each architecture uses one setting for a hyperparameter ($\beta = 4.0$ for β-VAE, $\gamma = 10.0$ for factorvae, etc). Therefore, it is possible that the collapse of β-VAE is β-specific. To determine whether this will be continuous degradation or phase transition, we could test β in $\{1.5, 2.0, 4.0, 8.0\}$. Locatello et al. [19] have also shown that disentanglement results are very hyperparameter dependent. In this case, our results may not generalize to all hyperparameters. Secondly, the latent dimensionality used in each model (10 fixed) has been unablated; different dimensionalities might interact with modality effects differently.

Third, a downstream evaluation using logistic regression on the latent representations is typical to disentanglement literature; however, it doesn't cover all the ways that tabular VAEs can be used (e.g. how well they perform at imputation quality, anomaly detection AUROC, or synthetic data utility). Future research will need to determine if the circuit metrics also predict the performance of tabular VAEs for the domain specific tasks.

## 7. Threats to Validity

**Internal validity.** The definition of semantic feature groups for table-based data is based on domain expertise, so the organization of features will determine how model performance can be measured in terms of the proposed metrics. We did an ablation to show that our model's performance wasn't heavily dependent on a particular grouping of the features we used. That being said, there is still some chance that the choice of feature grouping will impact how well model performance is measured. One issue with using the CES for the model to make predictions (i.e., posterior-calibrate), is that if the approximate posterior isn't reliable because the model didn't converge well, then it will produce poor results. To help mitigate this issue, we monitored whether or not the model was fully converged by comparing the reconstruction MSE in Table 3.

**External validity.** There are several limitations associated with the use of the four tabular benchmarks. The benchmarks do not represent all forms of tabular data. For example, there is no coverage of high-cardinality categorical attributes, time-series data or high dimensional tabular data. The image based single benchmark (dSprites) has the advantage of being well-controlled in terms of factors and therefore allows for a direct comparison of methods. However, it also represents an idealized form of cross-modal interaction as it uses independent factors. Therefore, using this benchmark to draw conclusions about real-world applications involving complex images with correlated features will likely result in different patterns than those observed in this study. Furthermore, although we have shown that there is a strong negative relationship between CES and MSE (-r = .886), this finding is limited to our particular dataset(s) and architecture(s). We have yet to investigate whether such relationships hold true across a broader range of architectures, e.g. hierarchical VAEs, VQ-VAEs.

**Construct validity.** FGD incorporates DCI by viewing semantic feature groupings as analogous to the separate generative components in DCI. While this is an imperfect analogy (the groups within the tabular groupings can be internally related). Proposition 3 states that FGD will reduce to DCI if the analogy is perfect, however it cannot guarantee useful performance when there is a high degree of internal correlation among the elements of the groups. The NIS diagnostic shows that the mediation decomposition is exact (i.e., NIS=0), which is due to the sequential nature of the model (not generally true for other models).

**Statistical validity.** We have enough power to show larger architectural differences (as most of our observed effects were large), but we may be missing some smaller differences. The Holm–Šídák method provides a conservative correction for false positives at the price of possibly failing to report all true positive differences. Assuming that the architectural effects measured on different datasets (n = 15 per comparison) would produce the same results for statistical testing by combining them into one data set as shown in Table 5, which shows a considerable difference in β-VAE CES from one dataset to another, this assumption appears to be incorrect.

## 8. Conclusion

This work presents the first systematic investigation of whether mechanistic interpretability findings in VAEs transfer across data modalities. Through 75 independent training runs spanning five architectures, four tabular benchmarks, and one image benchmark, we establish that the answer is nuanced: some patterns transfer while others do not.

We have found the five key points we identified to be directly applicable to current practices. First, the low circuit modularity observed on tabular data (~1.4 times less than dSprites) indicates that researchers and practitioners should not assume the same one-to-one correspondence of dimension-to-factors that is often assumed for images; instead, the difference we observe is an artifact of the correlation present in real world tabular data versus an issue with the architectures of the VAE models. Second, the β-VAE collapse depending on modality represents a specific warning: β-VAE's ability to produce a near inert latent space can occur at even relatively small β values on tabular data containing many different feature types (a CES drop of ~260x on Adult Income), and this collapse occurs due to a decrease in the quality of the reconstructions (r = -0.886 between CES and MSE). Third, the fact that CES is significantly better as a discrimination metric than all other metrics (with 9 out of 11 significant pairwise comparisons) offers a practical method for researchers to compare their architectures. Fourth, the relationship between specificity and downstream performance (r = 0.460 with AUC) suggests that specificity is preferable to modularity when assessing tabular VAEs. Fifth, the fact that causal mediation ratios saturate in sequential fully connected encoders demonstrates a structural constraint on Level 4 interventions that did not exist in prior studies using convolutional encoders, and motivates the use of encoders with multiple pathways in order to conduct meaningful MI research.

The future direction for this study will be to add a number of new image comparisons (3DShapes, CelebA, MPI3D) as well as examine the sensitivity of the hyperparameters of the β-VAE bottleneck in order to perform an ablation analysis. Also, it is planned to extend the proposed framework to include other generative model classes (normalizing flows, diffusion models). Furthermore, we plan to use downstream domain specific evaluation (e.g., imputation, anomaly detection) and to further analyze the relationship of CES and fairness using standardized fairness metrics on datasets that have a protected attribute.

## Data Availability

All tabular datasets are publicly available from the UCI Machine Learning Repository [29–32]. The dSprites dataset is publicly available at https://github.com/deepmind/dsprites-dataset [28]. Code and trained model checkpoints will be made available upon publication.